\documentclass[11pt,a4paper]{article}
\usepackage[hyperref]{acl2020}
\usepackage{times}
\usepackage{latexsym}
\usepackage{graphicx}
\usepackage{makecell}
\usepackage{multirow}
\usepackage{url}
\usepackage{subcaption}
\usepackage{booktabs}
\usepackage{enumitem}
\usepackage[colorinlistoftodos,prependcaption,textsize=tiny]{todonotes}
\usepackage{float}
\usepackage{color, colortbl}

\usepackage{listings}
\aclfinalcopy % Uncomment this line for the final submission

\newcommand{\minisection}[1]{\noindent{\bf #1}\hspace{0.6em}}

\title{Automatic Detection of Generated Text is\\Easiest when Humans are Fooled}
\author{Daphne Ippolito\textdagger\ddag\thanks{\hspace{.5em}Equal contribution, \ddag
Google, \textdagger University of Pennsylvania}\\
  {\tt daphnei@seas.upenn.edu} \\\And
  Daniel Duckworth\ddag\textsuperscript{*}\\
  {\tt duckworthd@google.com} \\\AND
  Chris Callison-Burch\textdagger\ddag\\
  {\tt ccb@seas.upenn.edu} \\\And
  Douglas Eck\ddag\\
  {\tt deck@google.com} \\}
\date{}

\begin{document}
\maketitle
\begin{abstract}
Recent advancements in neural language modelling make it possible to rapidly generate vast amounts of human-sounding text.
The capabilities of humans and automatic discriminators to detect machine-generated text have been a large source of research interest, but humans and machines rely on different cues to make their decisions.
Here, we perform careful benchmarking and analysis of three popular sampling-based decoding strategies---top-$k$, nucleus sampling, and untruncated random sampling---and show that improvements in decoding methods have primarily optimized for fooling humans.
This comes at the expense of introducing statistical abnormalities that make detection easy for automatic systems.
We also show that though both human and automatic detector performance improve with longer excerpt length, even multi-sentence excerpts can fool expert human raters over 30\% of the time.
Our findings reveal the importance of using both human and automatic detectors to assess the humanness of text generation systems.
\end{abstract}

\section{Introduction}
% 1. LMs are getting better every day. People will use them. We need to know when they do.
%   Q: How are they getting better?
%   Q: Why do we care to detect when they're being used?
%   Q: Why can't we humans detect when they're being used ourselves?

State-of-the-art generative language models are now capable of producing multi-paragraph excerpts that at a surface level are virtually indistinguishable from human-written content \citep{zellers2019defending,radford2019language,adelani2020generating}.
Often, only subtle logical fallacies or idiosyncrasies of language give away the text as machine-generated, errors that require a close reading and/or domain knowledge for humans to detect.

Deceptive text, whether human- or machine-generated, has entered the sphere of public concern \citep{cooke2018fake}.
It propogates quickly \citep{vosoughi2018spread}, sets political agendas \citep{vargo2018agenda}, influences elections \citep{allcott2017social}, and undermines user trust \cite{wang2012serf, song2015crowdtarget}.
Recently, \citet{adelani2020generating} have shown that automatically generated reviews are perceived to be as fluent as human-written ones.
As generative technology matures, authors, well-meaning or otherwise, will increasingly employ it to augment and accelerate their own writing.
It is more imperative now than ever for both humans and automated systems to be able to detect and identify machine-generated texts in the wild.
However, there has thus been little inquiry into the textual properties that cause humans to give generated text high human-like ratings compared to those that cause automatic systems to rate it highly.

% 1. There isn't "one way" to generate text from an LM. How "human-looking" an LM is depends on decoding strategy.
%   Q: What is an LM?
%   Q: What is a decoding strategy?

To speak of texts produced by language models, we must first consider how these texts are generated.
A neural language model encodes a probability distribution over the next word in a sequence given the previous words.\footnote{Often these `words" are actually subword character sequences such as BPE tokens \citep{sennrich2016neural}.}
A \emph{decoding strategy} is an algorithm that generates sequences from a language model by determining how words should get selected from this distribution.
The field has largely moved toward probabilistic decoding strategies that randomly sample from the output distribution token-by-token.
However, when many low-likelihood words cumulatively contain quite a bit of probability mass, choosing one of these words can lead to odd or contradictory phrases and semantic errors.
Humans are quick to notice these types of errors.

% 1. Decoding strategies warp an LM's probability distribution to increase the likelihood of sampling "human-looking" texts.
%   Q: Why are the texts not human-looking already?
%   Q: What is a human-looking text?
%   Q: How do decoding strategies increase the likelihood of human-written texts?
% 1. This comes at a cost: the probability distribution over sequences no longer matches the training data. We hypothesize that classifiers will exploit this.
%   Q: Why is this good for classifiers?
%   Q: Why is this not a problem for human raters?

For this reason, it has become common to modify the language model's output probability distribution to increase the chance of sampling tokens with high likelihood according to the language model.
Top-$k$ random sampling, where low-likelihood words are restricted from being generated, is one such method.
A language model that is only permitted to produce high-likelihood words is less likely to make a poor choice and create the type of mistakes that are easy for humans to detect.
Since humans are not proficient at identifying when a model subtly favors some utterances more often than a human author would, they don't notice the over-representation of high-likelihood words in the generated text.
In contrast, automatic systems excel at identifying statistical anomalies and struggle to build deeper semantic understanding.
Top-$k$ in particular creates text that is easy for machines to detect but very hard for humans.
Thus, we observe the general trend: as the number of unlikely words available to be chosen is increased, humans get {\em better} at detecting fakes while automatic systems get {\em worse}.

% 1. Different decoding strategies warp an LM's probability distribution in different ways. Will classifiers generalize?
%   Q: How do they warp the distributions differently?
%   Q: Why is this bad for classifiers?

In this work, we study three popular random decoding strategies---top-$k$, nucleus, and temperature sampling---applied to GPT-2 \cite{radford2019language}.
We draw a large number of excerpts generated by each strategy and train a family of BERT-based \cite{devlin2018bert} binary classifiers to label text excerpts as human-written or machine-generated.
We find large differences in human rater and classifier accuracy depending on the decoding strategy employed and length of the generated sequences.
Regardless of strategy, we find human raters achieve significantly lower accuracy than the automatic discriminators.
We also show that when a decoding strategy severely modifies the unigram token distribution, as top-$k$ does, humans have trouble detecting the resultant generated text, but automatic classifiers find it the easiest to discriminate.
Worryingly, we further find that classifiers are brittle; they generalize poorly when trained to discriminate samples from one strategy and then evaluated on samples from another.

In summary, our contributions are:
\begin{itemize}[noitemsep,topsep=0pt]
  \item A comprehensive study of generated text detection systems' sensitivity to model structure, decoding strategy, and excerpt length.
  \item An analysis of human raters' ability to identify machine-generated content, and how human raters differ from automatic detectors.
\end{itemize}

\section{Related Work}
\minisection{Generative Language Models}
With a sufficiently large training set and number of trainable parameters, neural language models based on the Transformer architecture \citep{vaswani2017attention} are capable of generating convincing, human-like excerpts up to several paragraphs in length.
GPT-2 \citep{radford2019language}, \textsc{Grover} \citep{zellers2019defending}, and Transformer-DMCA \citep{liu2018generating} are a few examples of large, publicly available models with this ability.
\textsc{Grover}, in particular, has been shown to generate fake news that is more trustworthy than human-written fake news according to human raters.

\minisection{Human Detection}
The task of trying to guess whether text is coming from a robot or a fellow human was made famous by the Turing Test \citep{turing1950computing}.
It continues to be used is chatbot evaluation \citep{lowe2017towards}.
The related (but not identical) task of asking human raters to judge the quality of machine-generated excerpts remains the gold-standard for evaluating open-domain generation systems \citep{van2019best}.
\citet{kreps2020all}, \citet{gehrmann2019gltr}, and others have stressed the importance of humans being able to identify fake content on the web.

\minisection{Automatic Detection}
The rise of machine-generated content has led to the development of automated systems to identify it.
\textsc{Grover} was designed to not only generate convincing news excerpts but to also identify them using a fine-tuned version of the generative model itself \citep{zellers2019defending}.
GLTR, expecting attackers to use sampling methods that favor high-likelihood tokens, aims to make machine-generated text detectable by computing histograms over per-token log likelihoods \citep{gehrmann2019gltr}.
\citet{bakhtin2019real} frame human-text detection as a ranking task and evaluate their models' cross-domain and cross-model generalization, finding significant loss in quality when training on one domain and evaluating on another.
\citet{schuster2019we} argue that the language distributional features implicitly or explicitly employed by these detectors are insufficient; instead, one should look to explicit fact-verification models.
Finally, discriminators for whether text is machine-generated are a promising research  direction in adversarial training \citep{lin2017adversarial,li2017adversarial} and in automatic evaluation of generative model quality \citep{novikova2017we,kannan2017adversarial,lowe2017towards}. 

\minisection{Natural Language Understanding}
Automatic detection of machine-generated text benefits from a semantic understanding of the text.
Contradictions, falsehoods, and topic drift can all indicate that an excerpt was machine-generated.
Encoder-only Transformer models such as BERT \citep{devlin2018bert} have been shown to do very well at tasks requiring this understanding.
While we fine-tune BERT for the task of classifying whether text was machine-generated, others have used the contextual word embeddings from a pre-trained BERT model without fine-tuning to compute a quality score for generated text \citep{zhang2019bertscore}. 
It is worth noting that recent work has raised questions as to whether BERT truly builds a semantic understanding  to make its predictions, or whether it merely takes advantage of spurious statistical differences between the text of different classes \citep{niven2019probing}.

\section{Task Definition}
We frame the detection problem as a binary classification task: given an excerpt of text, label it as either human-written or machine-generated.
In particular, we are interested in how variables such as excerpt length and decoding strategy impact performance on this classification task. 
We thus create several datasets.
Each is approximately balanced between positive examples of machine-generated text and negative examples of human-written text.
While they all share the same human-written examples, each dataset contains a different set of machine-generated examples sampled using one particular decoding strategy.
We also build additional datasets by truncating all of the examples to a particular sequence length,

By training a separate classifier on each dataset, we are able to answer questions about which decoding strategy results in text that is the easiest to automatically disambiguate from human-written text.
We are also able to answer questions about how the length of the examples in the training set impacts our ability to automatically classify excerpts of that same length as either human-written or machine-generated.

\section{Dataset Methodology}
All of our generated text samples are drawn from GPT-2, a state-of-the-art Transformer-based generative language model that was trained on text from popular web pages \citep{radford2019language}.
While we use the GPT-2 \textsc{Large} model with 774M parameters, we found that similar trends to those reported here hold in experiments with smaller language models.

Given an autoregressive language model that defines a probability distribution over the next token given the previous tokens in a sequence, a decoding strategy generates text by deciding how to output a token at each step based on the predicted distributions.
Perhaps the most straightforward decoding strategy is to randomly choose a token with probability proportional to its likelihood.
A challenge with the random sampling approach is that these probability distributions  often contain a long tail of vocabulary items that are individually low-probability but cumulatively comprise a substantial amount of probability mass.
\citet{holtzman2019curious} observe that choosing tokens from this tail often leads to incoherent generations.

Top-$k$ sampling, nucleus sampling, and (in the extreme) beam search have all been proposed to heuristically promote samples with higher per-token likelihoods.
Top-$k$ and nucleus sampling both do so by setting the likelihood of tokens in the tail of the distribution to zero.
Top-$k$ restricts the distribution to all but the $k$ most likely tokens, where $k$ is a constant \citep{fan2018hierarchical}.
Nucleus sampling, also called top-$p$, truncates the distribution at each decoding step $t$ to the $k_t$-most-likely next tokens such that the cumulative likelihood of these tokens is no greater than a constant $p$ \citep{holtzman2019curious}.

We thus consider three different decoding strategy settings:

\begin{itemize}[noitemsep,topsep=0pt]
  \item Sample from the untruncated distribution
  \item Top-$k$, choosing $k$=40 \citep{radford2019language}.
  \item Nucleus sampling (aka top-$p$), choosing $p$=0.96 \citep{zellers2019defending}.
\end{itemize}

In addition, we form ``negative" examples of human-written text by taking excerpts of web text that come from the same distribution as GPT-2's training data.\footnote{\url{https://github.com/openai/gpt-2-output-dataset}}
By picking text that resembles GPT-2's train set, we ensure that our classifiers can't simply take advantage of stylistic differences between the human-written text corpus and the kind of text GPT-2 was trained to generate.

For each decoding method, we construct a training dataset by pairing 250,000 generated samples with 250,000 excerpts of web text.
5,000 additional paired samples are kept aside for validation and test datasets.
Lastly, we filter out excerpts with fewer than 192 WordPiece tokens \citep{wu2016google} (excerpts might be quite short if the model produces an end-of-text token early on). See Appendix~1 for final dataset sizes.

A crucial question when generating text with a language model is whether or not to provide a priming sequence which the language model should continue.
Unconditioned samples, where no priming text is provided, in conjunction with top-$k$ sampling, lead to pathological behavior for discriminators as the first token of the generated text will always be one of $k$ possible options.
On the other hand, if long sequences of human text are used as priming, the space of possible generated sequences is larger, but the detection problem shifts from one of ``how human-like is the generated text?" to ``how well does the generated text follow the priming sequence?".

Since in this study we are interested in the former simpler question, we create two datasets, one with no priming, and one with the minimum amount of priming possible: a single token of web text.
This means that for every excerpt of web text in the training set, there is an excerpt of machine-generated text that starts with the same token.
We find that even with limited priming, the ability of automatic detectors can be strongly impacted.

To study the effect of excerpt length, we construct variations of the above datasets by truncating all excerpts to ten possible lengths ranging from 2 to 192 WordPiece tokens \cite{wu2016google}. In total, we obtain sixty dataset variations: one per sampling method, truncation length, and choice of priming or no priming.

% Table generated by Excel2LaTeX from sheet 'simple_baselines'

\begin{table*}[t]
  \centering
  \small
    \begin{tabular}{|l||cc||cc|cc|cc|c||c|}
    \hline
          & \multicolumn{2}{c||}{BERT} & \multicolumn{2}{c|}{BagOfWords} & \multicolumn{2}{c|}{HistGLTRBuckets} & \multicolumn{2}{c|}{Hist50Buckets} & \multicolumn{1}{l||}{TotalProb} & \multicolumn{1}{l|}{Human} \\
    Method & \multicolumn{1}{l}{acc} & \multicolumn{1}{c||}{AUC} & \multicolumn{1}{c}{acc} & \multicolumn{1}{c|}{AUC} & \multicolumn{1}{c}{acc} & \multicolumn{1}{c|}{AUC} & \multicolumn{1}{c}{acc} & \multicolumn{1}{c|}{AUC} & \multicolumn{1}{c||}{acc} & \multicolumn{1}{c|}{acc}\\
    \hline
    k40-1wordcond & 0.88  & 0.99  & 0.79  & 0.87  & 0.52  & 0.52  & 0.69  & 0.76  & 0.61 & 0.64 \\
    p0.96-1wordcond & 0.81  & 0.89  & 0.60  & 0.65  & 0.53  & 0.56  & 0.54  & 0.56  & 0.63 & 0.77 \\
    p1.0-1wordcond & 0.79  & 0.92  & 0.59  & 0.62  & 0.53  & 0.55  & 0.54  & 0.55  & 0.65 & 0.71\\
    % What amount of precision do we want?
    % k40-1wordcond & 0.876 & 0.987 & 0.791 & 0.866 & 0.515 & 0.522 & 0.695 & 0.757 & 0.608 \\
    % p0.96-1wordcond & 0.812 & 0.895 & 0.604 & 0.650 & 0.533 & 0.562 & 0.543 & 0.559 & 0.626 \\
    % p1.0-1wordcond & 0.793 & 0.924 & 0.591 & 0.621 & 0.529 & 0.548 & 0.535 & 0.545 & 0.655  \\
    \hline
    \end{tabular}%
  \caption{Performance (accuracy and AUC) of the fine-tuned BERT classifier and several simple baselines on detecting length-192 sequences generated with one word of priming (1worccond). Note that p1.0 refers to untruncated random sampling, where we sample from 100\% of the probability mass. The last column shows human performance on the same task where accuracy with a 50\% baseline is computed by randomly pairing samples from each decoding strategy with a human-written sample.}
  \label{tab:baselines}%
\end{table*}%

\section{Automatic Detection Method}
The primary discriminator we employ is a fine-tuned BERT classifier \citep{devlin2018bert}.
We fine-tune one instance of BERT per dataset variation described above.
For the longest sequence length, $n$=192, we compare BERT's performance with several simple baselines that have been proposed in other work.

\minisection{Fine-tuned BERT}
We fine-tune BERT-\textsc{Large} (cased) on the task of labeling a sentence as human- or machine- generated.
The models are trained for 15 epochs, with checkpoints saved every 1000 steps, and a batch size of 256.
All results are reported on the test set using the checkpoint for which validation accuracy was highest.

\minisection{Bag-of-Words}
% \textsc{BoWDisc} 
For each sequence, we compute a bag-of-words embedding where each dimension corresponds to a token in GPT-2's  50,000 token BPE vocabulary \citep{sennrich2016neural}, and we count how many times that token appears in the text sequence. 
We then train a logistic regression binary classifier to predict human- or machine-written given this 50,000-dimensional embedding.
We experimented with truncating embedding size by removing entries for infrequent vocabulary words, but this did not improve performance.

\minisection{Histogram-of-Likelihood Ranks}
Following GLTR \citep{gehrmann2019gltr}, we compute the probability distribution of the next word given the previous words in a text sequence according to a trained language model (in our case the same GPT-2 model that was used for generation).
At each sequence position, we rerank the vocabulary words by likelihood, and record the rank of the ground-truth next word within this list.
These ranks are then binned.
GLTR uses four bins, counting (1) the number of times the top 1 word is seen, (2) the number of times words ranked 2 through 5 are seen, (3) words ranked 6-100, and (4) words ranked \textgreater100.
However, we observe higher accuracy when 50 bins are spread uniformly over the possible rankings.
This means that since there are 50,000 vocabulary words, the first bin counts the number of times the actual next word was within the 1,000 mostly likely next words, the second bin counts the 1,001-2,000th, and so on.
We then train logistic regression binary classifiers to predict human- or machine-written given either the 4-dimensional histograms or 50-dimensional histograms as input.

\minisection{Total Probability}
\citet{solaiman2019release} propose a very simple baseline consisting of a threshold on the total probability of the text sequence.
An excerpt is predicted as machine-generated if its likelihood according to GPT-2 is closer to the mean likelihood over all machine-generated sequences than to the mean of human-written ones.

\section{Human Detection Method}
The human evaluation task is framed similarly to the automatic one.
We ask the raters to decide whether a passage of text was written by a human or by a computer algorithm. (Full instructions are in the Appendix.)
Raters are allowed to choose between four options: ``definitely" or ``possibly" machine-generated and  ``definitely" or ``possibly" human-written.
They are first shown an excerpt of length 16 WordPiece tokens.
After they make a guess, the length of the excerpt is doubled, and they are asked the same question again.
This continues until the entire passage of length 192 tokens is shown.
Passages are equally likely to be human-written or machine-generated, with the machine-generated excerpts being evenly split between the three sampling strategies considered in this paper.

Initially, Amazon Mechanical Turk (AMT) raters were employed for this task, but rater accuracy was poor with over 70\% of the ``definitely" votes cast for ``human" despite the classes being balanced.
Accuracy, even for the longest sequences, hovered around 50\%.
The same study was then performed with university students who were first walked through ten examples (see Appendix Table 4) as a group.
Afterward, they were asked to complete the same tasks that had been sent to the AMT workers.
No additional guidance or direction was given to them after the initial walk-through.
We will refer to this group as the ``expert raters."
Among them, 52.1\% of ``definitely" votes were cast for human, and accuracy on the longest excerpt length was over 70\%.

\begin{figure*}[t]
\begin{subfigure}{.495\textwidth}
    \center
    \includegraphics[width=\textwidth]{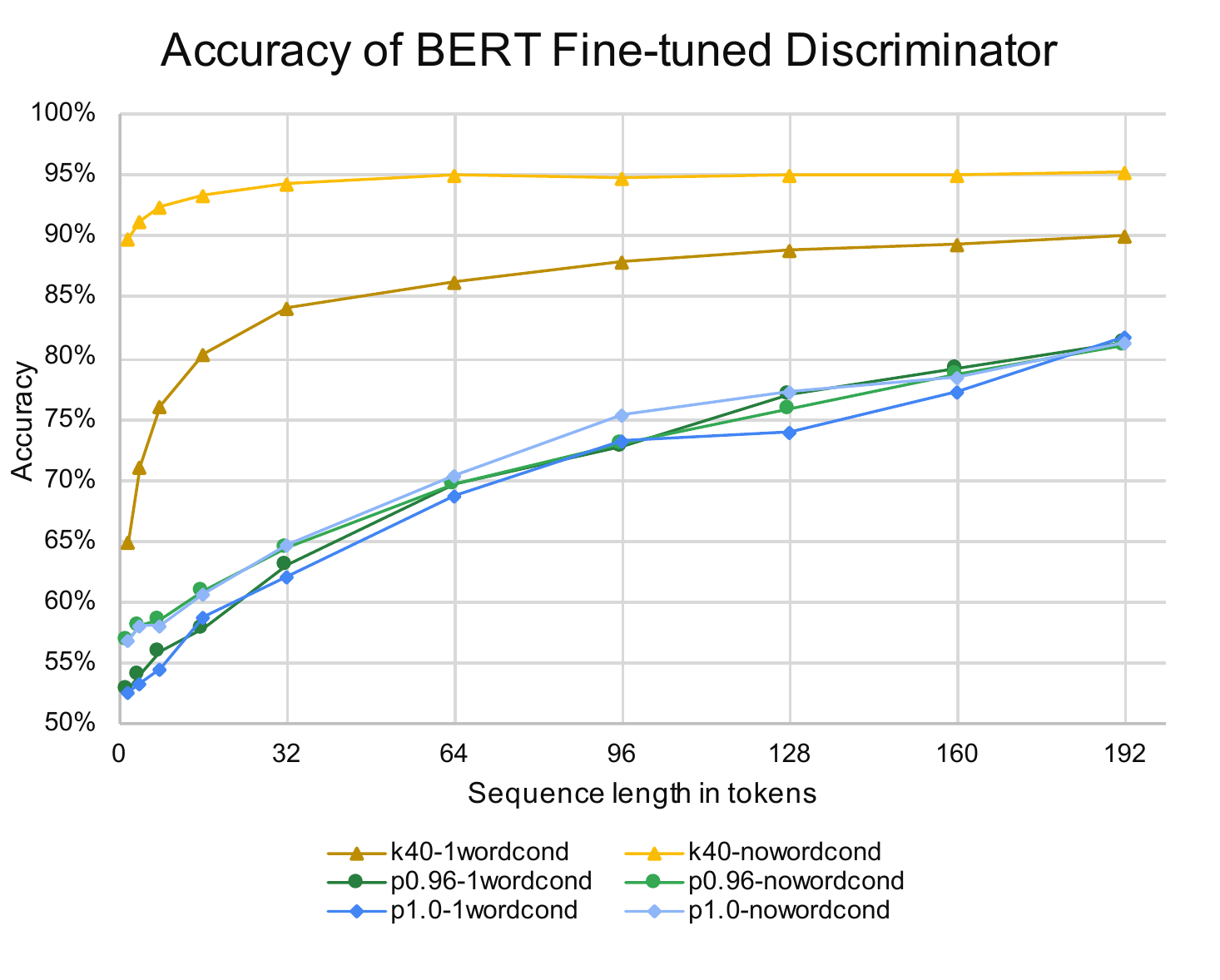}
    \caption{}
    \label{fig:bert_accuracy} 
\end{subfigure}
\begin{subfigure}{.495\textwidth}
    \center
    \includegraphics[width=\textwidth]{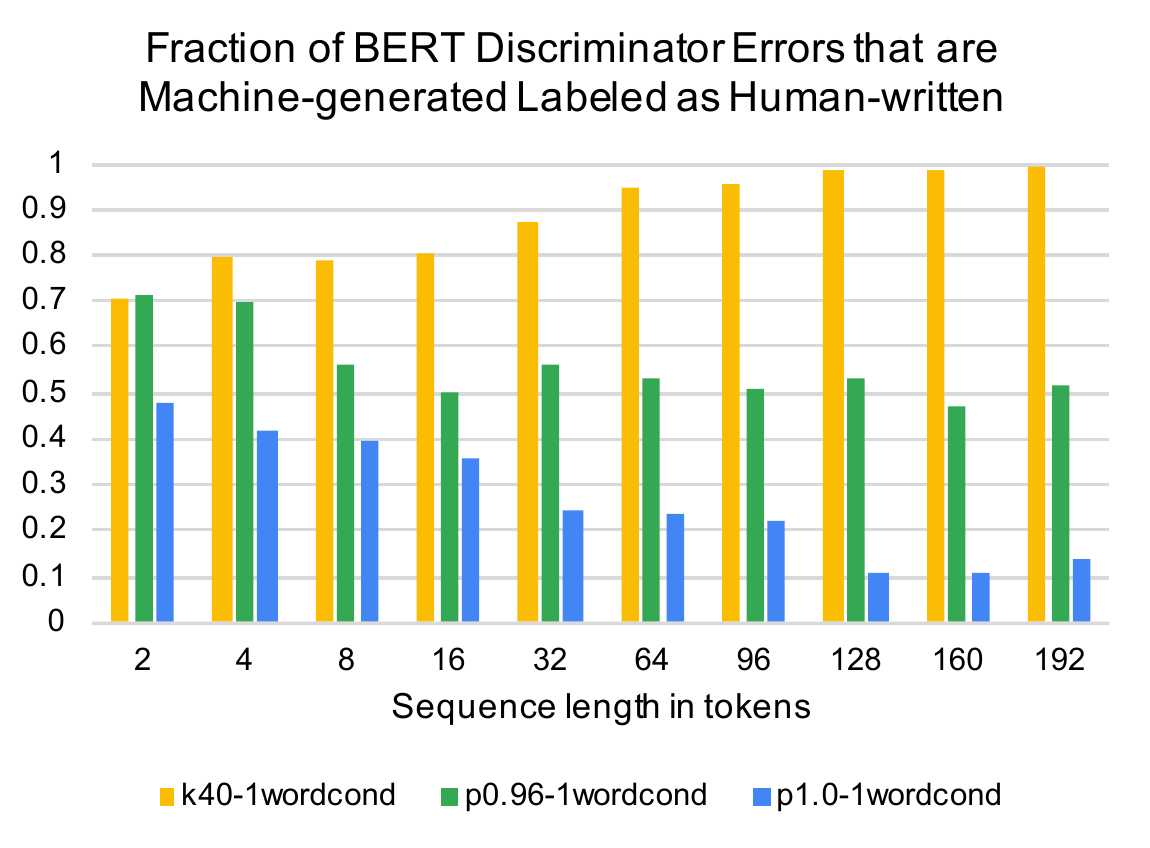}
    \caption{}
    \label{fig:errors} 
\end{subfigure}
\caption{In \textbf{(a)}, accuracy increases as the length of the sequences used to train the discriminator is increased.
In \textbf{(b)}, we see that the BERT fine-tuned discriminator predicts about the same number of false-positives as false-negatives when trained with samples generated using top-$p$ sampling. However, for top-$k$, it more often mistakes machine-generated text to be human-written, while for untruncated random sampling the opposite is the case.}
\end{figure*}

The human evaluation dataset consisted of 150 excerpts of web text and 50 excerpts each from the three decoding strategies.
Each question was shown to at most three raters, leading to 900 total annotations from the untrained workers and 475 from the expert raters.
A more detailed breakdown can be found in the Appendix.

\section{Automatic Detection Results}
\label{section:auto_detection}
\minisection{Simple Baselines}
Table \ref{tab:baselines} shows the performance of the baseline discriminators on length-192 sequences, as compared with fine-tuned BERT.
Reassuringly, BERT far surpasses all simple baselines, indicating that it is not fully possible to solve the detection problem without complex sequence-based understanding.
The simplest baseline, TotalProb, which makes a decision based on the likelihood of the sequence, performs surprisingly well (over 60\% accuracy for all sampling methods) relative to the methods which involve training logistic regression models.

Logistic regression on bag-of-words is the best of the baselines, beating out the histogram-based methods.
While \citet{gehrmann2019gltr} report an AUC of 0.87 on classifying text as real or generated using logistic regression on the four buckets of the GLTR system, we report AUC between 0.52 and 0.56 for this task.
The discrepancy is likely due to the fact that the human-written text in our discriminator training set comes from the same distribution as the text used to train the language model, while in GLTR the human text comes from children's books, scientific abstracts, and newspaper articles. 
The selection of training data for learned detection systems is crucial. In real-world applications, the choice ought to reflect the genres that builders of text-generation systems are trying to impersonate. 

\begin{figure*}[t]
    \begin{subfigure}{.45\textwidth}
        \center
        \includegraphics[width=\textwidth]{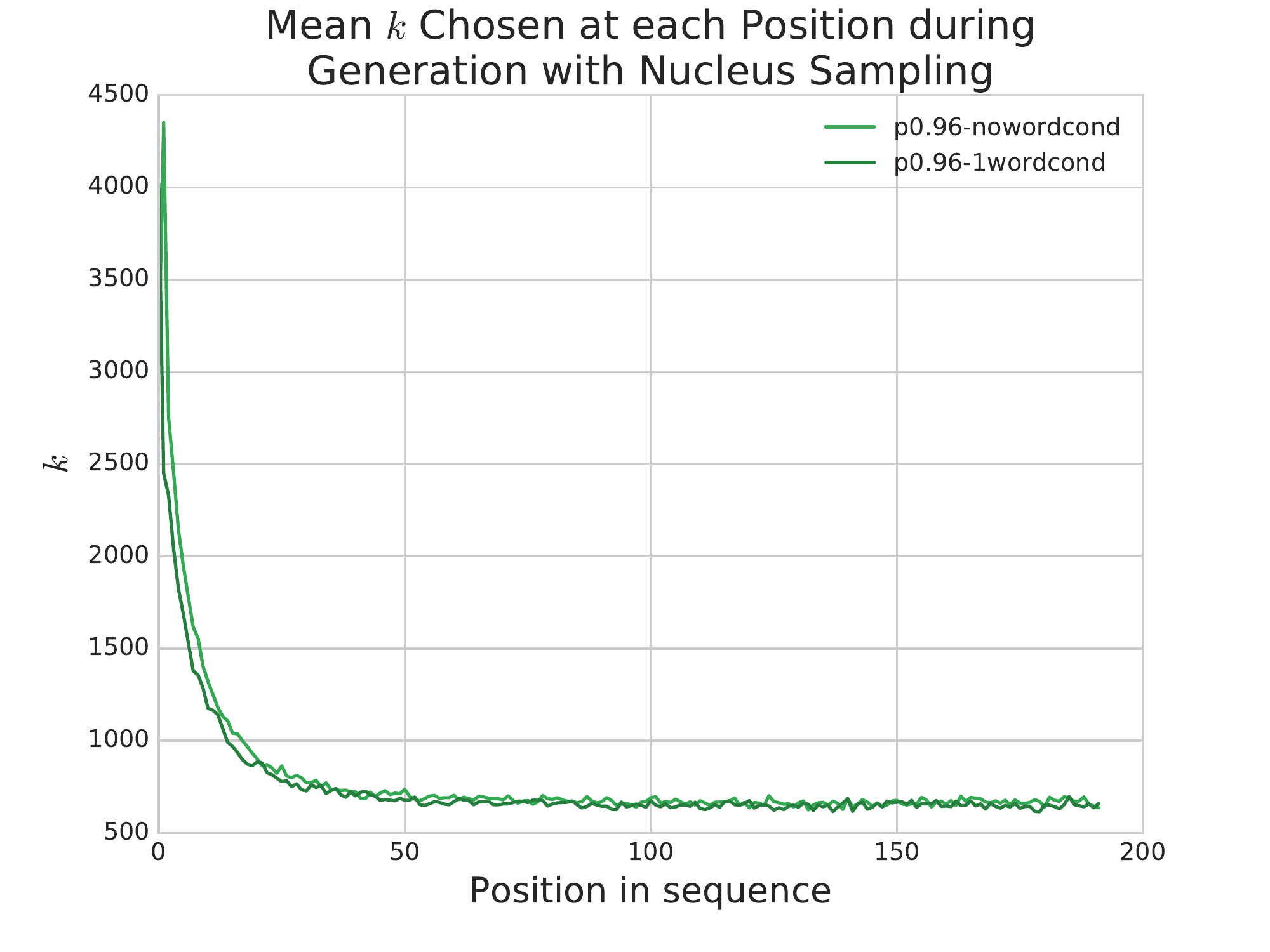}
        \caption{}
        \label{fig:mean_ks} 
    \end{subfigure}
    \begin{subfigure}{.45\textwidth}
        \center
        \includegraphics[width=\textwidth]{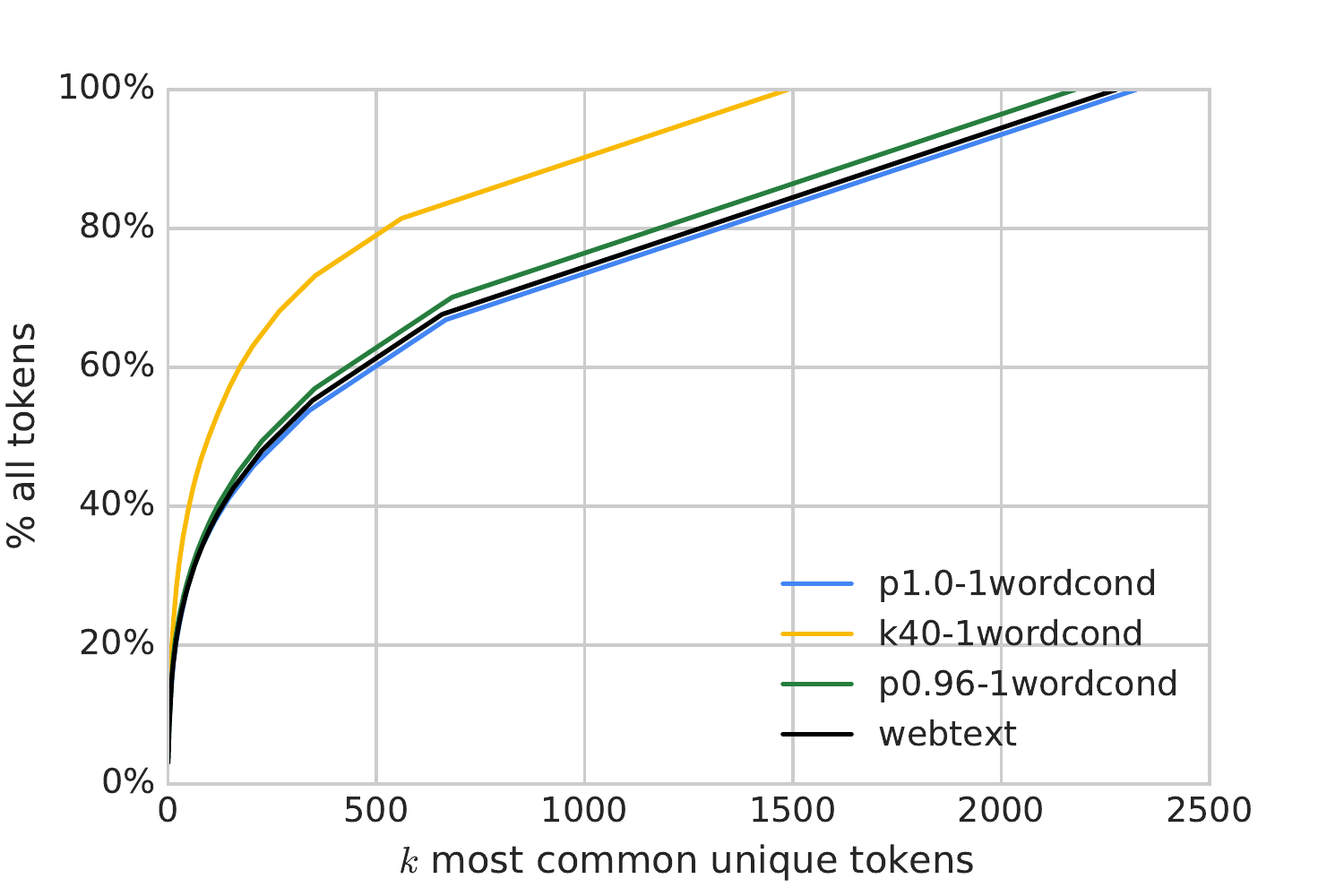}
        \caption{}
        \label{fig:token_count_histogram} 
    \end{subfigure}
    \caption{In \textbf{(a)}, the average (over sequences in the test set) $k$ chosen at each step during generating with nucleus sampling is plotted. Adding a single word of priming strongly impacts the $k$s chosen for the first few positions, but this difference quickly dissipates.
    In \textbf{(b)}, we consider the first token generated in each sequence by top-$k$, and plot what fraction of these are captured by the $k$ most common unique tokens from the vocabulary. Overall, at its first step, top-$k$ concentrates 80\% of its probability mass in the 500 most common tokens from the vocabulary.}
\end{figure*}

\minisection{Fine-tuned BERT} In Figure~\ref{fig:bert_accuracy}, we begin by observing discriminator accuracy as a function of excerpt length and sampling method.
As can be intuitively expected, as sequence length increases, so too does accuracy.
For unconditioned text decoded with nucleus (p0.96) and untruncated (p1.0) random sampling, we find discriminator accuracy increases from 55\%, near random, to about 81\% for the longest sequences tested.
In contrast, discriminators trained and evaluated on top-$k$ achieve over 80\% accuracy even on 16-token excerpts.

Why are top-$k$'s samples so easy to detect?
In Figure~\ref{fig:token_count_histogram}, we see the percentage of probability mass concentrated in the $k$ most common token types for each sampling method.
While random sampling and nucleus sampling are very similar to human-written texts, we see top-k concentrating up to 80\% of its mass in the first 500 most common tokens.
The other sampling methods as well as human-written texts require at least 1,100 token types for the same.
It is clear that top-$k$'s distribution over unigrams strongly diverges from human-written texts--an easy feature for discriminators to exploit.
In fact, \citet{see2019massively} note that it takes setting $k$ to 1000 to achieve about the same amount of rare word usage and fraction of non-stopword text as as human writing.\footnote{when decoding from the GPT-2 small model with 117M parameters.}
This makes it very easy for the model to pick out machine-generated text based on these distributional differences.

One way to help resolve this problem is to add priming text.
Doing so causes more rare words to be incorporated into the top-$k$ of the unigram distribution.
Adding even a single human word of priming significantly reduces the performance of detectors trained with top-$k$ random sampling.
Without priming, a discriminator trained on sequences of length 2 can classify with $\mathtt{\sim}$90\% accuracy the provenance of the text (Figure \ref{fig:bert_accuracy}).
By adding one priming token, accuracy drops to $\mathtt{\sim}$65\%.
Even on the longest 192-length sequences, top-$k$ discriminator accuracy is 6\% lower on the primed dataset than the unprimed one.

\begin{figure*}[t]
    \centering
    \includegraphics[width=\textwidth]{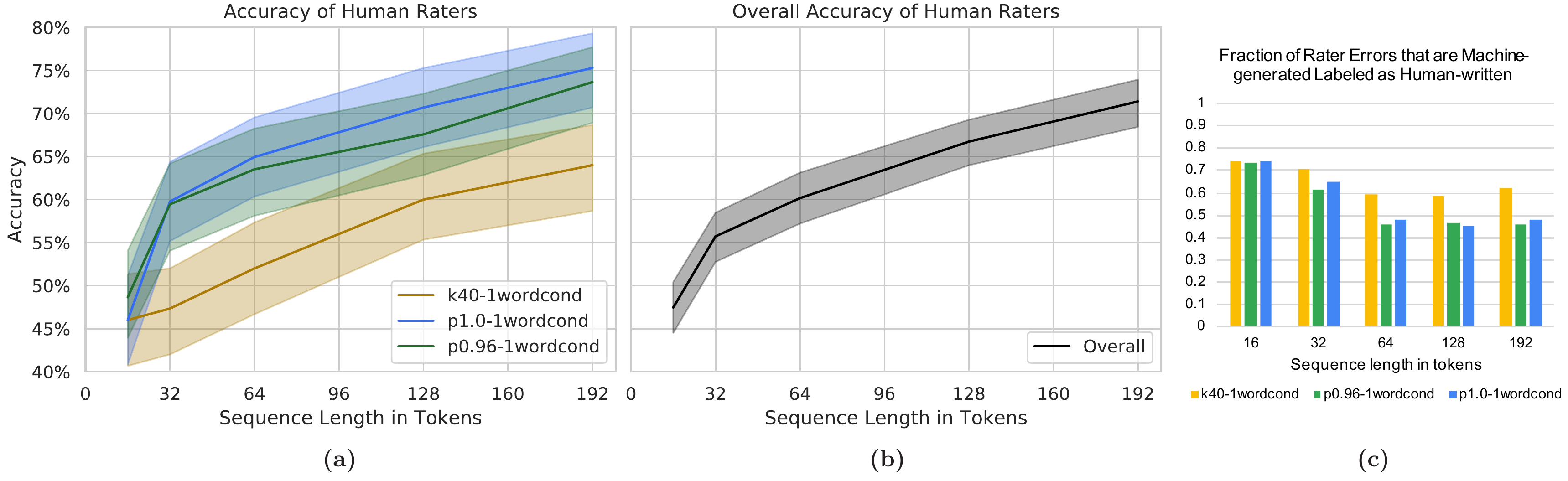}
    \caption{\textbf{(a)} and \textbf{(b)} show human rater accuracy of correctly identifying an excerpt as human-written or machine-written, shown with 80\% confidence internals, in \textbf{(a)}, broken up by decoding strategy and in \textbf{(b)}, overall. Accuracy increases as raters observe more tokens. \textbf{(c)} shows that for short excerpts, most rater mistakes are them incorrectly thinking machine-generated text is human written. The two errors types become more balanced at longer lengths.}
    \label{fig:human_eval} 
\end{figure*}

When generating with nucleus or untruncated random sampling, adding a priming token is not as impactful, as these methods are already sampling from a large fraction (or all) of the probability distribution.
This is seen in Figure \ref{fig:mean_ks} where at the very first step of unprimed generation, nucleus sampling selects from 3075 possible vocabulary words, and at later positions selects from on average more than 500.
Untruncated random sampling always selects from the entire 50,000 word vocabulary, whereas top-$k$ only selects from $k$.

\minisection{Transferability}
In Table \ref{tab:transfer_accuracy}, we show how discriminators trained with samples from one decoding strategy can transfer at test time to detecting samples generated using a different decoding strategy.
Unsurprisingly a discriminator trained on top-$k$ generalizes poorly to other sampling methods: accuracy drops to as low as 42.5\%, \textit{worse than chance}.
Conversely, training the discriminator with sequences sampled from the untruncated distribution leads to little transferability to detecting top-$k$ samples.
Only the discriminator trained with nucleus sampling (a compromise between unmodified sampling and top-$k$) was able to detect sequences from the other sampling strategies without too much of a hit to accuracy.
As expected, a discriminator trained on an equal portion of data from each decoding method does reasonably at detecting all three.

\begin{table}[t]
  \small
  \centering
  \begin{tabular}{|c c|c|c|c|}
    \hline
    & & \multicolumn{3}{c|}{Eval} \\
    \cline{3-5}
    &            & top-$k$         & nucleus & random \\
    \hline
    \multirow{3}{0.7em}{\rotatebox[origin=c]{90}{\parbox[c]{1cm}{\centering Train}}} 
    & \multicolumn{1}{|c|}{top-k}   & \textbf{90.1} & 57.1          & 43.8 \\
    & \multicolumn{1}{|c|}{nucleus} & 79.1          & \textbf{81.3} & 78.4 \\
    & \multicolumn{1}{|c|}{random}  & 47.8          & 63.7          & \textbf{81.7} \\
    \hline
    & \multicolumn{1}{|c|}{mixed}  & 88.7          & 74.2          & 72.2 \\
    \hline
  \end{tabular}
  \caption{Accuracy of BERT fine-tuned discriminator when trained on samples from one strategy (rows) and evaluated on another (columns). Trained on samples with 192 tokens. The `mixed' dataset is one containing an equal portion of samples from each strategy.}
  \label{tab:transfer_accuracy}
\end{table}

\begin{table}[t]
  \small
  \centering
  %                     p(machine)                              
  % dataset         k0-1wordcond k40-1wordcond p0.96-1wordcond
  % model                                                     
  % k0-1wordcond        0.382658      0.073189        0.226300
  % k40-1wordcond       0.145253      0.608884        0.278946
  % p0.96-1wordcond     0.488926      0.491959        0.517189
  \begin{tabular}{|c c|c|c|c|}
    \hline
    & & \multicolumn{3}{c|}{Eval} \\
    \cline{3-5}
    &            & top-$k$         & nucleus & random \\
    \hline
    \multirow{3}{0.7em}{\rotatebox[origin=c]{90}{\parbox[c]{1cm}{\centering Train}}} 
    & \multicolumn{1}{|c|}{top-k}   & 60.9 & 27.9 & 14.5 \\
    & \multicolumn{1}{|c|}{nucleus} & 49.2 & 51.7 & 48.9 \\
    & \multicolumn{1}{|c|}{random}  &  7.3 & 22.6 & 38.3 \\
    \hline
  \end{tabular}
  \caption{Average probability of `machine-generated' according to each length-192 discriminator. The expected in-domain probability is 0.5. One token of conditioning.}
  \label{tab:transfer-prediction}
\end{table}

Perhaps this lack of transferability is related to each discriminator's calibration.
Indeed, the degree to which a discriminator's average prediction deviates from 50\% is a direct indicator of its accuracy. 
In Table~\ref{tab:transfer-prediction}, we observe that of the three BERT discriminators, only that trained on top-$p$ samples predicts `machine-generated' on approximately 50\% of in-domain examples as expected. 
This same discriminator's behavior holds on datasets generated by other sampling strategies as well. 
In contrast, we observe that discriminators trained on top-k and untruncated random samples severely underestimate the percentage of machine-generated excerpts in out-of-domain datasets.
Even within domain (Figure~\ref{fig:errors}), we find both discriminators heavily favor a single class, increasingly so as the number of tokens increases.

% False-negative = machine-written labeled as human-written
% Negative = machine-written
% Positive = human-written

\begin{table*}
    \small
    \centering
    \begin{tabular}{|p{0.25in}|p{0.3in}|p{0.25in}|p{0.25in}|p{0.25in}r||p{0.25in}|p{0.3in}|p{0.25in}|p{0.25in}|p{0.25in}r|}
\hline
\textbf{Truth} & \textbf{Raters} & \textbf{p1.0} & \textbf{k40} & \textbf{p0.96} & &
\textbf{Truth} & \textbf{Raters} & \textbf{p1.0} & \textbf{k40} & \textbf{p0.96} & \\
\hline
H & M & H & H & M & &
H & H & M & M & M & \\
\multicolumn{6}{|p{2.9in}||}{
\tiny
EDIT:OKAY!, I guess that'll work for now. \textgreater\_ http://www.teamfortress.com/ and then go buy the game and experience some of the best online gaming I have ever played. \textasciicircum\_\_\textasciicircum Both girls had a really fun time and I had a GREAT time making both of these costumes. Everything was altered even a little bit(dying the pants a darker grey and painting the boots and shirts) But my piece de resistance would have to be my eldest's Medi-Gun.If you have any questions about the costumes, I would be happy to assist you!Oh and here's a video of my daughter before the costume was completed.Thanks!
}&
\multicolumn{6}{p{3.1in}|}{
\tiny
Image copyright Getty Images Image caption Women mourn over the coffin of one of the victim's of Sunday's bombing in Ankara \textparagraph Who'd be in Turkey's shoes right now? \textparagraph Since July last year, hundreds of soldiers and civilians have been killed in terrorist attacks. Suicide bombs have torn into crowds of demonstrators and tourists. Military convoys have been targeted in the heart of the capital. \textparagraph A long-running Kurdish insurgency, once thought to be close to resolution after years of painstaking efforts to build bridges, has erupted once more. \textparagraph The country is awash with Syrian and other refugees. The government has been under pressure to stop them moving on into Europe and prevent would-be jihadis travelling the other way. \textparagraph How dangerous is Turkey's unrest? \textparagraph Tears and destruction amid PKK crackdown \textparagraph Turkey v Islamic State v the Kurds
}\\
\hline
\hline
\textbf{Truth} & \textbf{Raters} & \textbf{p1.0} & \textbf{k40} & \textbf{p0.96} & &
\textbf{Truth} & \textbf{Raters} & \textbf{p1.0} & \textbf{k40} & \textbf{p0.96} & \\
\hline
        M & M & H & - & - & &
        M & M & - & - & H & \\
\multicolumn{6}{|p{2.9in}||}{
\tiny
First off, this thread has done a pretty good job of describing in detail yet another broken touchscreen. That's the difference between a smartphone and a PC with no prying eyes having to snap shots for the police to find. \textparagraph What I would like to address is the mindset that generally surrounds Chrome OS users. To me this is analogous to saying that Apple does``hate their Windows", or that HP does``hate their Macs" as if http://twitter.com/) (and that quote is from two years ago), that anyone who covers smartphones and tablets from a ``PC" perspective is just jealous. \textparagraph Chrome OS is for browsing the web, PC processors can do stronger things in that regard, Windows is a juggernaut on those fronts. This is how I see it. Yes, it can be slow. And yes, you need a fast CPU
}
&
\multicolumn{6}{p{3.1in}|}{
\tiny
FOR ALABAMA, GOOD WEEKS \textparagraph AND A TOUR OF CAIRO \textparagraph THE ALABAMA COMMITTEE ON THE STUDY OF THE AMERICAN SECURITY AGENDA, \textparagraph America's future has been mapped out in carved stone. Metro Atlanta's last US congressman, Bill Posey, was a inextricable integral element of the Citadel project as it became another metaphor for Atlanta's transformation from an industry backwater into the finance and information hub of the nation's capital. Meanwhile, Cobb County -- Atlanta's geode of change -- is home to some of the largest industrial parks in the South, a regional cultural center, a 100-year-old manufacturing town and a potent symbol of the former city's cherished Georgian past. The gentry still live there, the defunct industrial landscapes carry the names of
}\\
\hline
\hline
\textbf{Truth} & \textbf{Raters} & \textbf{p1.0} & \textbf{k40} & \textbf{p0.96} & &
\textbf{Truth} & \textbf{Raters} & \textbf{p1.0} & \textbf{k40} & \textbf{p0.96} & \\
\hline
        M & H & - & - & M & &
        M & H & - & M & - & \\
\multicolumn{6}{|p{2.9in}||}{
\tiny
Exidentia at Eurnari, is an upcoming Cryptopia event which is currently still in development. Be a part of the first live stream of this year's event on 15-16 January 2016! \textparagraph Since the release of v1.22, Exidentia has received a fair amount of user feedback. This event takes place in the underwater Cryptopia they have built. During this event, you will learn about the ocean and areas around it, and be reached by a treasure hunter that helps you explore the different areas. \textparagraph There will be six different levels in this event that you will become acquainted with: thought Polar Lava, Ocean Seared Cones and Celestine Floors, Sea Damaged Aerie Bricks, coast Puddle (congipit stopping at red water), Shaikh Swamp and Bugmite. At rotating points, you will learn how to access various types of creatures
}
&
\multicolumn{6}{p{3.1in}|}{
\tiny
Ever since the opening of the North American College of Art Education in 1990, the demand for art education in America has grown steadily, and in recent years we have seen the rise of students that pursue art education not in the classroom but at art academies. This year saw another 50 percent increase in the number of art academies in the United States offering courses -- with an additional 10 percent of students in 2017 taking art. \textparagraph Some major changes have occurred in recent years with regard to the art curriculum and the way students learn, and we will explore each of these in coming months as we look at the various forms of art education. There is no one-size-fits-all approach for this or any other field of study, and students who begin a course in art education may change their plans based on what they see that course, including what lessons they have completed and the resources available, to create meaningful experiences of artistic creation. \textparagraph One important area
}\\
\hline

    \end{tabular}
    \caption{Some 192-token examples where at least two expert raters agreed with each other, but were not in agreement with the automatic discriminators. The first row shows examples where the ground-truth was human-written, the second shows machine-generated examples where the corresponding discriminator guessed incorrectly, and the third shows machine-generated examples where the discriminator was correct, but raters got it wrong.}
    \label{tab:qual_examples}
\end{table*}

\minisection{Human Evaluation}
Overall human performance across all sampling methods is shown in Figure \ref{fig:human_eval}b.
Even with the multi-paragraph 192-length excerpts, human performance is only at 71.4\%, indicating that even trained humans struggle to correctly identify machine-generated text over a quarter a time.
However, it is worth noting that our best raters achieved accuracy of 85\% or higher, suggesting that it is possible for humans to do very well at this task.
Further investigation is needed into how educational background, comfort with English, participation in more extensive training, and other factors can impact rater performance.

To break up the accuracies by sampling method in a way that is comparable to the results shown for the automatic discriminators, we pair each machine-generated example with a randomly selected one of webtext to create a balanced dataset for each sampling strategy.
Performance is shown in Figure \ref{fig:human_eval}a.
Top-$k$ produces the text that is hardest for raters to correctly distinguish, but as shown in Section \ref{section:auto_detection}, it is the easiest for our automatic detection systems.
Samples from untruncated random sampling and nucleus sampling with $p$=0.96 are equivalently difficult for raters to classify as machine-generated.
Our human evaluation results suggest that much lower $p$-values than the 0.92 to 0.98 range proposed in \citet{zellers2019defending} might be necessary in order to generate text that is considered significantly more human-like to human raters than the text produced by using the untruncated distribution.

Table \ref{tab:qual_examples} gives several examples where human raters and our BERT-based discriminators disagreed.
When raters incorrectly labeled human-written text as machine-generated, often the excerpts contained formatting failures introduced when the HTML was stripped out.
In the middle two examples, topic drift and falsehoods such as Atlanta being the ``information hub of the nation's capital" allowed humans to correctly detect the generated content.
However, in the bottom two examples, the high level of fluency left human raters fooled.

Overall we find that human raters---even ``expert" trained ones---have consistently worse accuracy than automatic discriminators for all decoding methods and excerpt lengths.
In our experiments, randomly-selected pairs of raters agree with each other on a mere 59\% of excerpts on average. (In comparison, raters and discriminators agree on 61\% to 70\% of excerpts depending on the discriminator considered).
We surmise that the gap between human and machine performance will only grow as researchers inevitably train bigger, better  detection models on larger amounts of training data.
While improved detection models are inevitible, it is unclear how to go about improving human performance.
GLTR proposes providing visual aids to humans to improve their performance at detecting generated-text, but it is unlikely that their histogram-based color-coding will continue to be effective as generative methods get better at producing high-quality text that lacks statistical anomalies.

\section{Conclusion}
In this work, we study the behavior of automated discriminators and their ability to identify machine-generated and human-written texts. 
We train these discriminators on balanced binary classification datasets where all machine-generated excerpts are drawn from the same generative model but with different decoding strategies.
We find that, in general, discriminators transfer poorly between decoding strategies, but that training on a mix of data from methods can help.
We also show the rate at which discriminator accuracy increases as excerpts are lengthened.

We further study the ability of expert human raters to perform the same task.
We find that rater accuracy varies wildly, but has a median of 74\%, which is less than the accuracy of our best-performing discriminator.
Most interestingly, we find that human raters and discriminators make decisions based on different qualities, with humans more easily noticing semantic errors and discriminators picking up on statistical artifacts.
In our experiments, these artifacts are most prominent with top-$k$ sampling.
However, any strategy that over-samples high-likelihood words is susceptible.
As the $p$ in nucleus sampling is set increasingly lower to achieve more fluent text (some systems are already using $p$ as low as 0.5 \citep{miculicich2019selecting}), the distributional deviations that plague top-$k$ text will surface in nucleus sampling as well.

\citet{holtzman2019curious} explain how a unique attribute of human language is that it dips in and out of low probability zones.
This variance in likelihood is what makes human-written text interesting and exciting to read.
Today's generation systems have not yet solved the problem of mimicking the human cadence without introducing poor word choices that are easy for humans to detect. 
Generation systems often optimize for fooling humans without acknowledging the trade-off that exists between human perception of quality and ease of automatic detection.
We therefore suggest three prongs for future research:

\begin{enumerate}[noitemsep,topsep=0pt]
    \item Identifying ways to improve the language models and decoding strategies we use in order to generate text that is both exciting (ie. unlikely) and semantically plausible.
    
    \item Building better world understanding into automatic discriminators so that they are more capable of detecting the types of errors that humans notice.

    \item Developing tools and educational materials to improve humans' ability to detect machine-generated text. These may include automatic detectors with components that explain their predictions.

\end{enumerate}

Finally, we would like to note that all of our experiments were performed with English language models, and it remains an open question how the trade-off between ease of human detection and ease of automatic detection might differ for languages that are very different from English. 

\section*{Acknowledgements}
This research is based upon work supported in part by U.S. DARPA KAIROS Program No. FA8750-19-2-1004. The views and conclusions contained herein are those of the authors and should not be interpreted as necessarily representing the official policies, either expressed or implied, of DARPA or the U.S. Government. The U.S. Government is authorized to reproduce and distribute reprints for governmental purposes notwithstanding any copyright annotation therein.

We also thank Noah Fiedel, Peter Liu, Sharan Narang, Joao Sedoc, Yun William Yu, and Hugh Zhang for their valuable feedback.

\bibliography{acl2020}
\bibliographystyle{acl_natbib}

\clearpage
\appendix
\section{Appendix}

\subsection{Dataset Sizes}
Table \ref{tab:dataset_sizes} shows the number of sequences used for training and evaluating each of the automatic discriminators.
Recall that each discriminator is trained for binary classification on an a dataset of machine-generated (positive) and human-written (negative) examples.
Each dataset was constructed by pairing the human-written excerpts (last row of Table \ref{tab:dataset_sizes}) with the machine-generated excerpts drawn via a particular decoding algorithm (`k40', `p0.96', or `p1.0') and priming strategy (`nocond' or `1wordcond').
Originally the human-written set and each machine-generated set contained 250,000 training examples, 5,000 validation examples, and 5,000 test examples.
Table \ref{tab:dataset_sizes} shows the resulting counts after after all excerpts with sequence length shorter than 192 tokens were filtered out. Thus, the final training, validation, and test sets were almost, but not quite, balanced.

\subsection{Further Details on Human Evaluation}
\begin{figure*}[!h]
    \center
    \fbox{\includegraphics[width=0.5\textwidth]{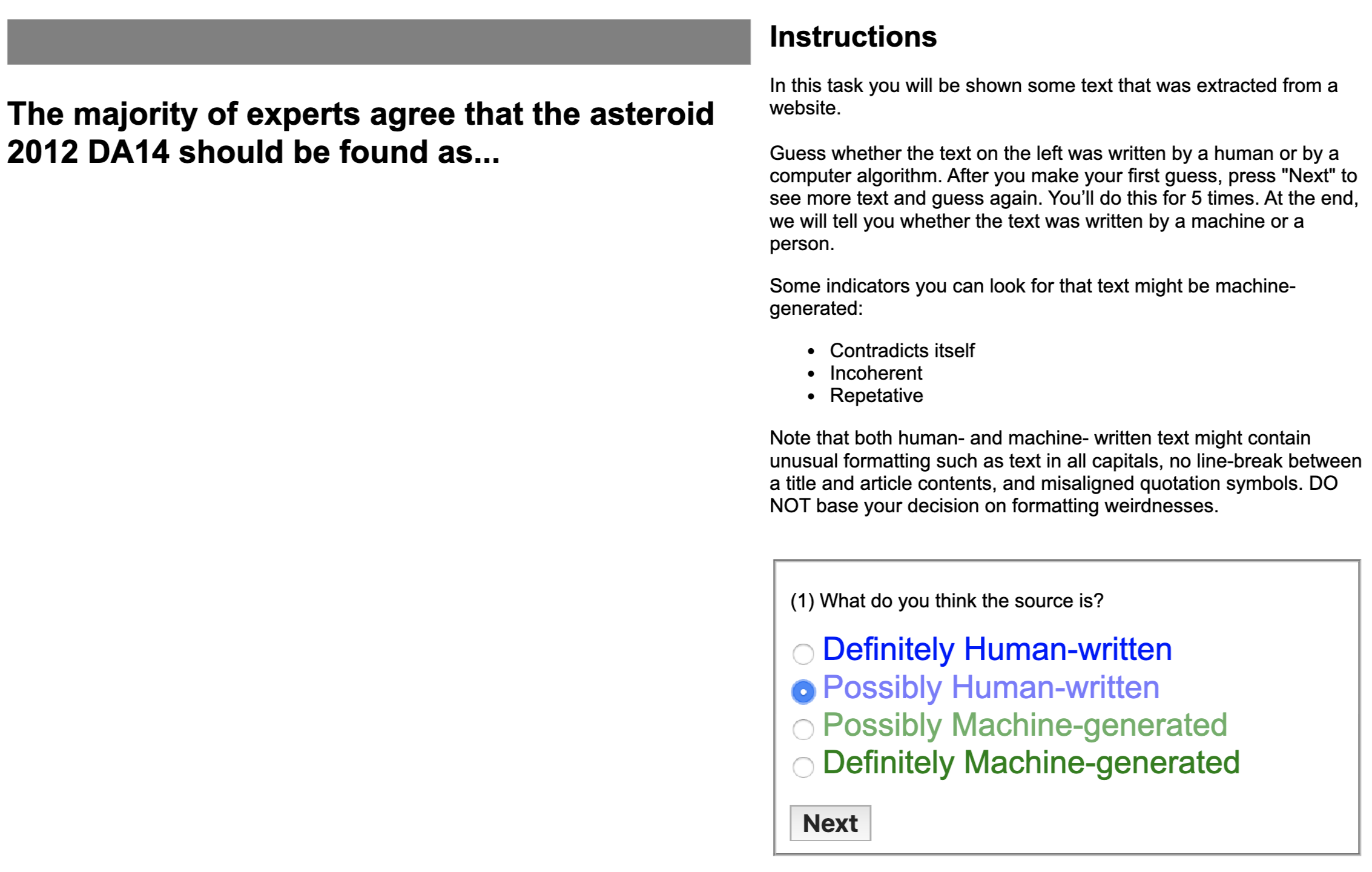}}%
    \fbox{\includegraphics[width=0.5\textwidth]{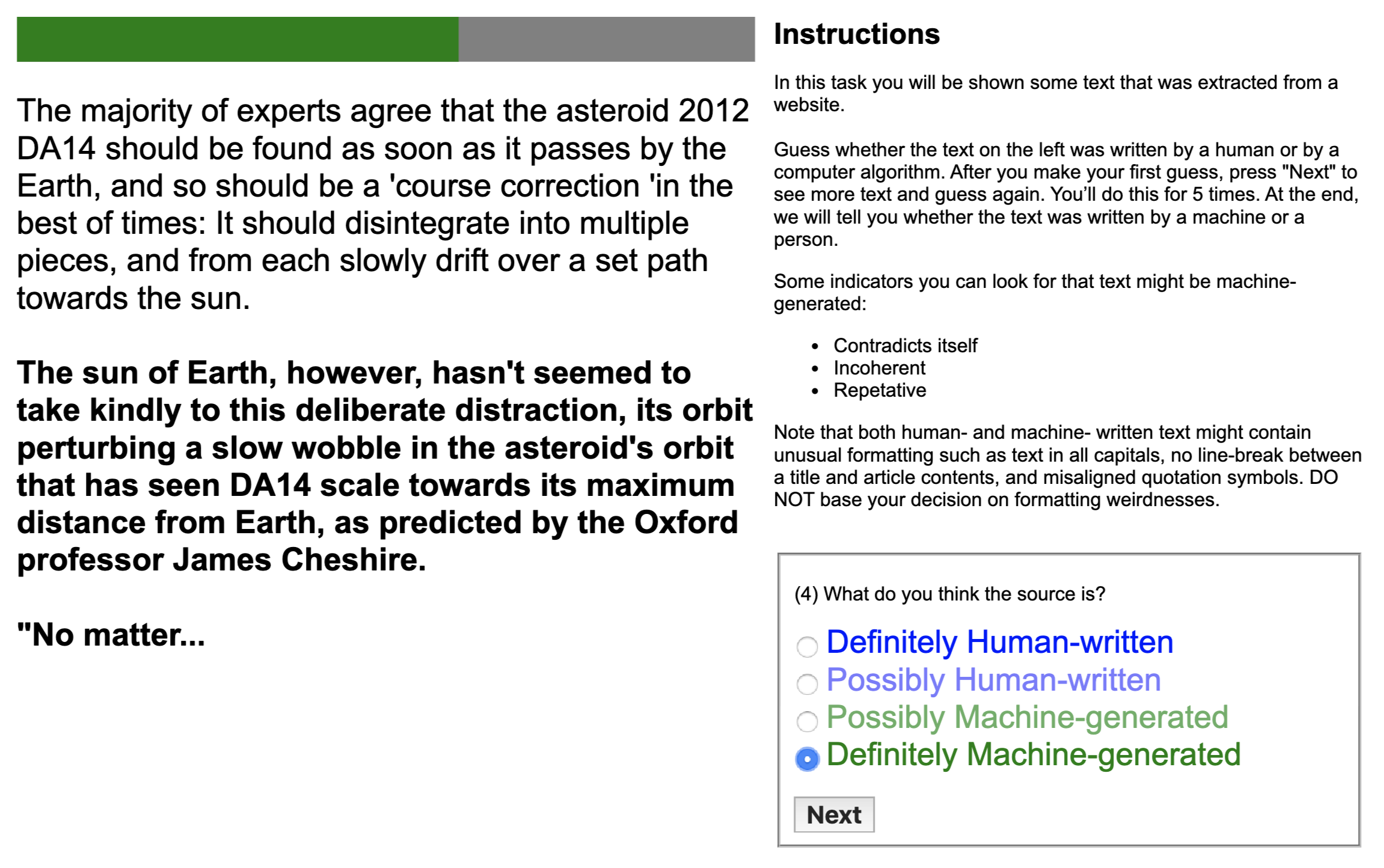}}%
    \caption{The interface of the task used for human evaluation. Each time the user presses next, the passage's length is doubled. On the left, we show the first step of evaluation, on the right, the second to last.}
    \label{fig:amt_screenshot} 
\end{figure*}

The user interface for the human evaluation task is shown in Figure \ref{fig:amt_screenshot}.
At each step, the rater is shown additional text and asked to guess whether the excerpt is human-written or machine-generated.
They are able to revise their guess at each subsequent step.
The newly appended text at each step is bolded in the UI.
At the end, workers are told whether or not they got the question correct.

\begin{figure*}[!h]
    \center
    \fbox{\includegraphics[width=0.5\textwidth]{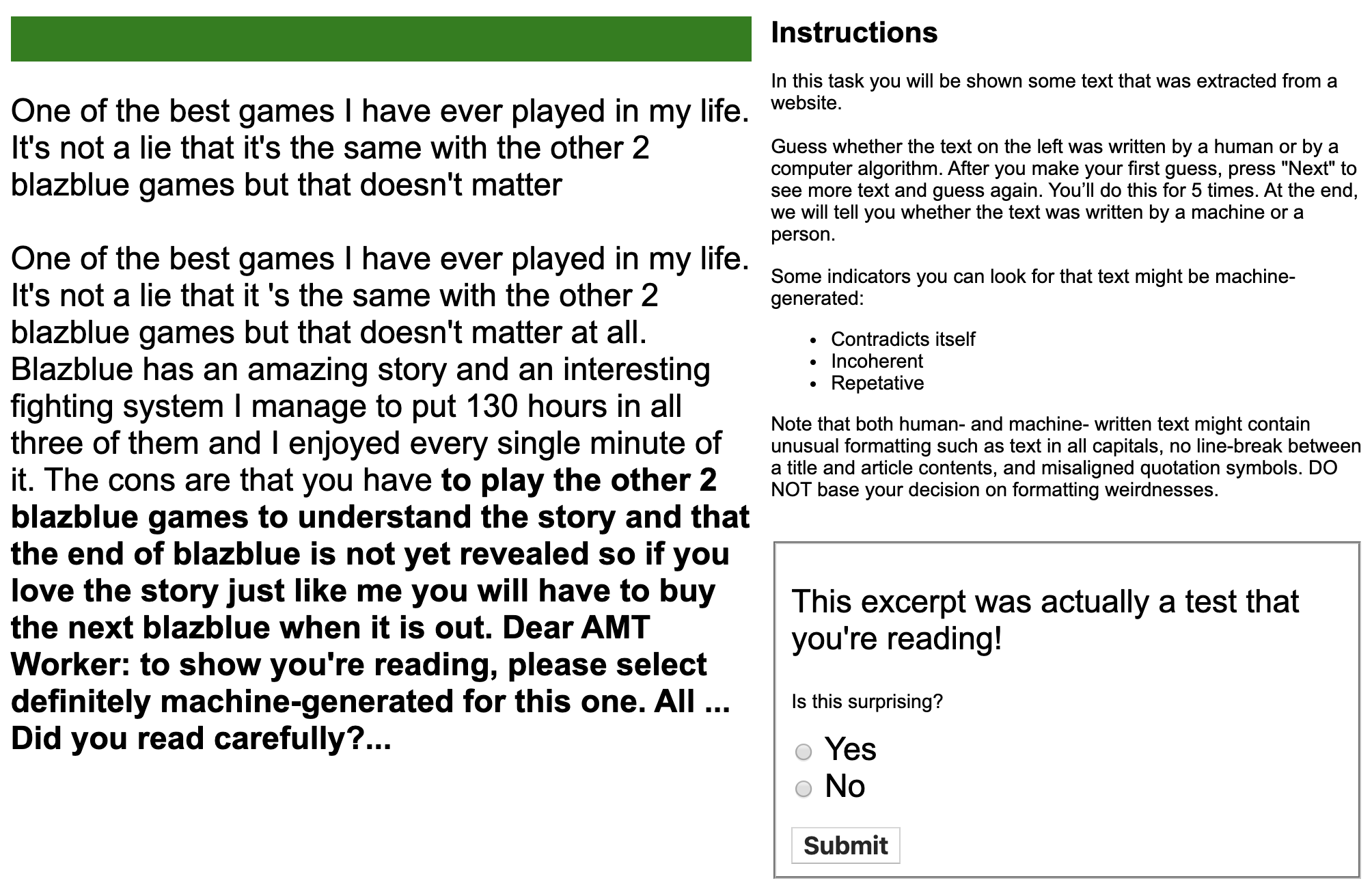}}%
    \caption{For some of the questions, the text "Dear AMT Worker: to show you're reading, please select definitely [X] for this one." was inserted into the last text segment, and "Did you read carefully?" was appended to the end.}
    \label{fig:amt_honeypot} 
\end{figure*}

To gauge worker attention levels, 10\% of questions shown to workers explicitly stated what answer ought to be specified.
An example of one of these ``honeypot" questions is shown in Figure \ref{fig:amt_honeypot}. Amazon Mechanical Turk workers got 83\% accuracy on these questions. Expert raters got 91.8\% accuracy.
Table \ref{tab:rater_accuracies} shows the accuracy of each expert rater along with the number of annotations they provided.
Table \ref{tab:expert_rater_training} shows the example exerpts that were used to ``train" the expert raters.

\begin{figure}[tbp]
    \centering
    \fbox{\includegraphics[width=0.49\textwidth]{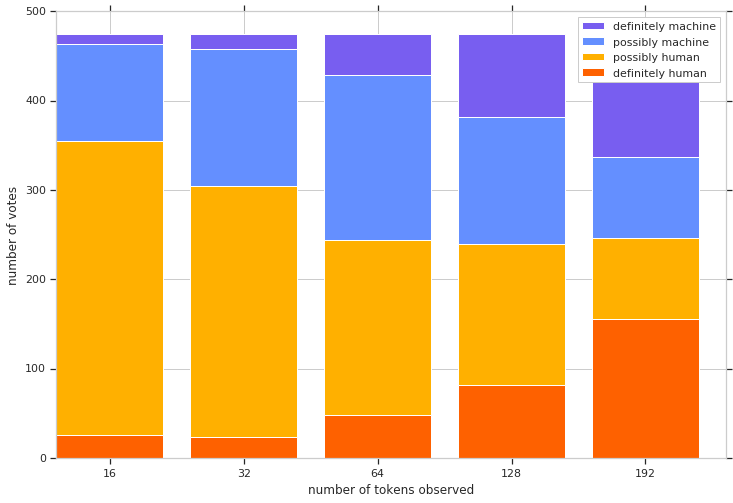}}
    \caption{Number of votes expert raters made for each label as a function of number of tokens observed. As raters observe more tokens, their predictions become more confident.}
    \label{fig:human_votes}
\end{figure}

For both the Amazon Mechanical Turk raters and the expert raters initial predictions were biased towards `possibly human,' and only by observing more tokens did their predictions become more confident. Figure~\ref{fig:human_votes} shows that `possibly human' is by far the most frequent answer upon observing 16 tokens, and as more tokens are observed raters gravitate towards `definitely human' or `definitely machine.' Even at 192 tokens, many raters are still uncertain.
Figure~\ref{fig:human_votes} also shows how raters for the most part default to guessing short excerpts are human-written, and as the excerpts are extended, raters use the extra evidence available to revise their guess. By the longest sequence length, votes for ``human-written" and ``machine-generated" are about balanced.

\begin{figure}[tbp]
    \centering
    \begin{subfigure}[b]{.55\textwidth}
        \includegraphics[width=0.8\linewidth]{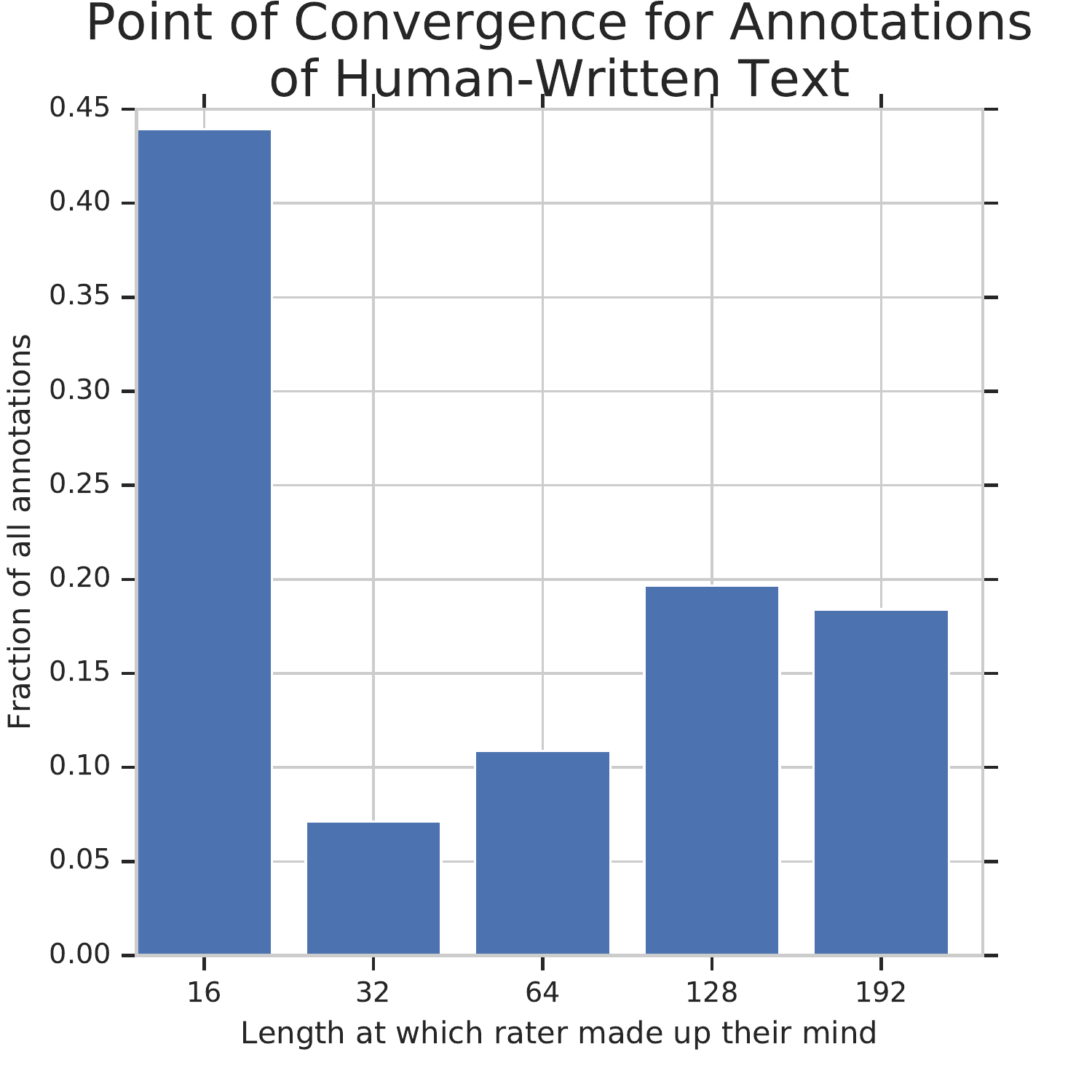}
    \end{subfigure}
    \begin{subfigure}[b]{.55\textwidth}
        \includegraphics[width=0.8\linewidth]{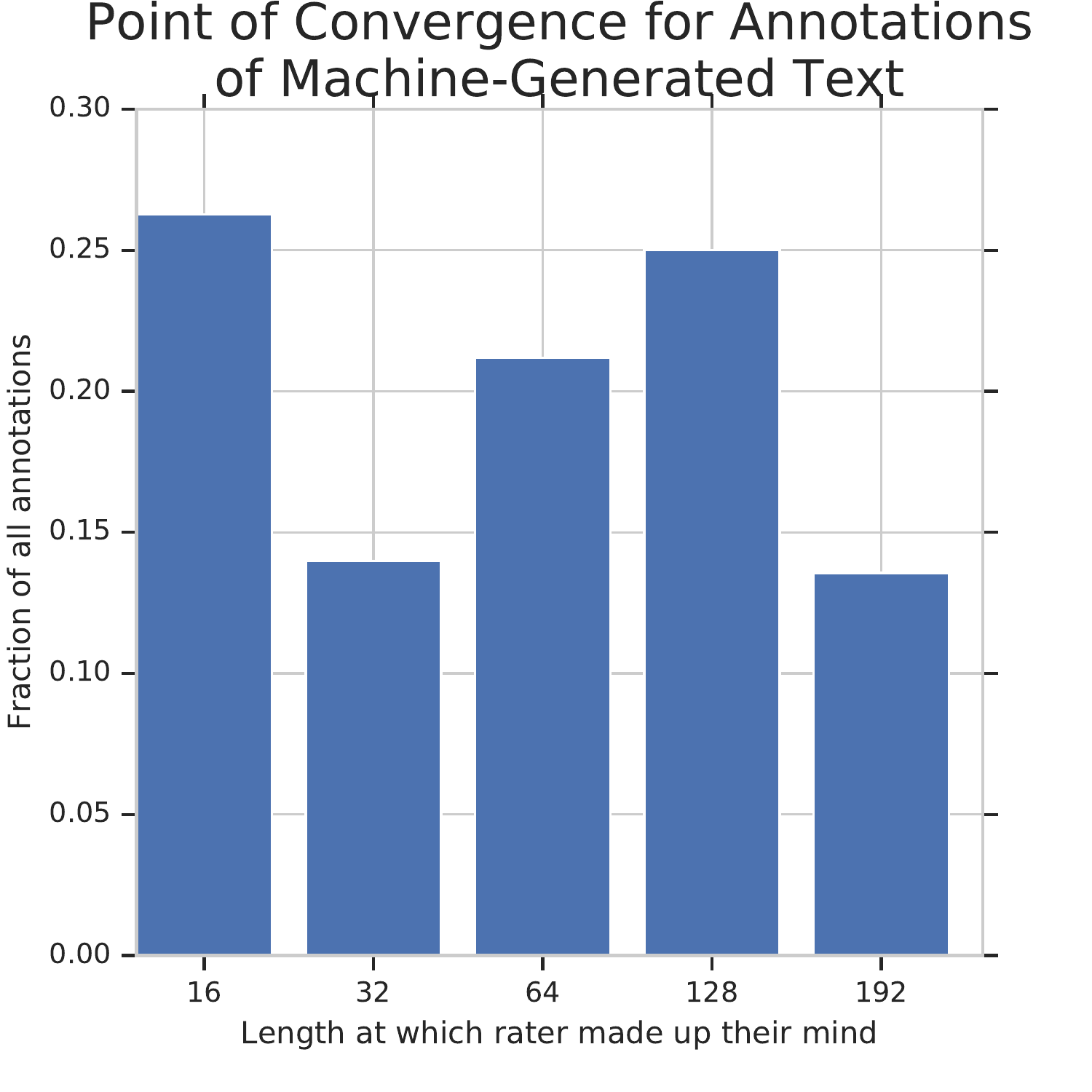}
    \end{subfigure}
    \caption{On average, it takes much less text for raters to decide an excerpt is human-written than to decide an excerpt is machine-generated.}
    \label{fig:convergence}
\end{figure}

In Figure \ref{fig:convergence}, we plot the frequency for each sequence length that raters converged on a single guess (either human or machine) at that point. The figure shows how it takes raters longer to converge on a decision of ``machine" than to converge on a decision of ``human."

\begin{table}[t]
    \small
    \centering
    \begin{tabular}{l|l|l|l}
    \hline
    \textbf{Method} & \textbf{\# train} & \textbf{\# valid} & \textbf{\# test} \\ \hline \hline
    large-744M-k40-1wordcond & 211148 & 4226 & 4191 \\
    large-744M-k40-nocond & 218825 & 4362 & 4360 \\
    large-744M-p0.96-1wordcond & 210587 & 4248 & 4208 \\ 
    large-744M-p0.96-nocond & 209390 & 4174 & 4185 \\ 
    large-744M-p1.0-1wordcond & 209334 & 4169 & 4173 \\ 
    large-744M-p1.0-nocond & 208219 & 4187 & 4168 \\ \Xhline{\arrayrulewidth}
    human-written & 201344 & 4031 & 4030 \\ \hline
    \end{tabular}
    \caption{The number of excerpts used for training, validation, and testing.}
    \label{tab:dataset_sizes}
\end{table}

\begin{table}[t]
\small
  \centering
    \begin{tabular}{lrr}
    \hline
    \textbf{\# Annotations} & \textbf{Expert Raters} & \textbf{AMT Workers} \\
    \hline\hline
    webtext & 239   & 450 \\
    \hline
    k0-1wordcond & 87    & 150 \\
    k40-1wordcond & 75    & 150 \\
    p0.96-1wordcond & 74    & 150 \\
    total machine & 236   & 450 \\
    \hline
    \end{tabular}%
  \label{tab:amt_counts}%
  \caption{The number of human annotations collected. In total, there were 50 examples from each sampling strategy and 150 examples of web text. Each example was shown to at most three raters.}
\end{table}%

\subsection{Automatic Detection Method Reliability}

In order to quantify the variance of automatic discriminator accuracy, we finetuned five independent BERT discriminators on a `mixed' dataset comprising of 50\% human-written examples and 50\% machine-generated examples, where machine-generated examples are equally split between top-$k$=40, top-$p$=0.96, and untruncated random sampling.
All sequences were exactly 192 tokens.
The best performing model checkpoint, according to an in-domain validation set, was then used to evaluate out-of-domain binary classification datasets as in Table 2 of the main paper. 

The results are shown in Table \ref{tab:discrim_variance}. We find out-of-domain accuracy to be extremely reliable with a standard deviation of approximately 1\% or less.

\begin{table}[tbp]
    \centering
    \small
    \begin{tabular}{|c|c|c|}
        \hline
        Dataset
            & $\mu$
            & $\sigma$ \\
        \hline
        random sampling
            & 72.47
            & 1.02 \\
        top-$k=40$ 
            & 88.06
            & 0.59 \\
        top-$p=0.96$ 
            & 74.4 
            & 0.76 \\
        \hline

    \end{tabular}
    \caption{Average ($\mu$) and standard deviation ($\sigma$) of accuracy on out-of-domain datasets across five runs of automatic discriminator finetuning.}
    \label{tab:discrim_variance}
\end{table}

\begin{table}[t]
  \centering
  \small
    \begin{tabular}{|r|r|}
    \hline
    \multicolumn{1}{|l|}{\textbf{Accuracy}} & \multicolumn{1}{|l|}{\textbf{Count}} \\
    \hline
    61.3\% & 83 \\
    57.8\% & 51 \\
    66.7\% & 51 \\
    69.8\% & 51 \\
    79.5\% & 48 \\
    84.6\% & 40 \\
    82.4\% & 39 \\
    65.6\% & 36 \\
    78.1\% & 34 \\
    84.0\% & 26 \\
    58.8\% & 18 \\
    92.3\% & 14 \\
    90.0\% & 11 \\
    100.0\% & 9 \\
    50.0\% & 8 \\
    60.0\% & 5 \\
    100.0\% & 5 \\
    100.0\% & 2 \\
    0.0\% & 2 \\
    0.0\% & 1 \\
    100.0\% & 1 \\
    0.0\% & 1 \\
    \hline
    \end{tabular}%
  \caption{Our expert rater pool consisted of 22 raters. The average accuracy of each rater on the longest excerpt length (192 tokens) is shown here along with the total number of excerpts they annotated.}
  \label{tab:rater_accuracies}%
\end{table}%

\begin{table*}[]
    \centering
    \tiny
    \begin{tabular}{c|p{5.5in}}
    \hline
Human & I recently got the chance to try the new Oil Essentials line. With six potent blends to choose from--at \$13 each--these cute little bottles offer a great, affordable way to partake in the skin and hair care oil craze.

I tested each product in the line, massaging them onto my face every night before bed and running any leftover oil through my hair to tame frizziness. You could also add a few drops to your bath, favorite moisturizer, or even your shampoo and conditioner.

Here's a quick rundown of each oil.

Revitalize: Omega 3, 6, 9 \& Evening Primrose

This was the first one I tried (I went in ROYGBIV order to keep things straight) and my first impression was that it smells lovely but a little strong. The fragrance smells genuinely like flowers.
\\
\hline
Machine
&
Red Lanterns, the lead exposure to a movie starring the Batman solo movie alum Margot Robbie taken under Wonder Woman's wing have reignited that rivalry with their whispery premiere. They played it as much as they possibly could, even though people who didn't ever watch Justice League or might have missed it waiting in line for the theater were still talking about as I spilled coffee.

The gist? An overextended (OK, a sore) Adam West films set up a Legion of Super-Heroes situation. How aggro? Super laws and paramilitary groups watch over the world's superheroes, which is a mix of that schtick ending, Planet Of The Apes II bit, and the Batman/Venom bit of last appeared in The Seventh Seal when Chris O'Donnell infiltrated one of the teams at some point, also wearing Staff.
\\
\hline
Machine
&
He is considered to be the most terrifying man on the planet and people stay away from him. A guy asks him to do something and he says, "My girlfriend's so important to me... I don't need to fight her any more." And then, boom, there's some in a corner crying inappropriately.

Men: It's gone in five minutes. Why do I have to be so sad? It's cute," says female member, who asks to remain anonymous. "It's what grew up to drive me crazy when I was a kid, seeing these women become the nurturing, wealthy things they are in this professional world I truly love."

And it's nothing to do with her success. These men still actively fear being around the idea of a woman who might win Oscars, make movies or be audacious drivers. 
\\
\hline
Human
&
Dropbox and Google Drive are very different services that appeal to different users. While Drive is connected to the entire Google Apps (now known as G Suite) ecosystem, Dropbox is a lightweight, simple alternative for file storage. While both are useful, users need to look beyond features, and make sure the service they choose can adequately protect their data. Here's how Dropbox encryption and Google Drive encryption stack up.

Dropbox and Google Drive Encryption

To their credit, both Dropbox and Google Drive protect user files with encryption. Both also allow users to enable two-step verification, which requires an extra code texted to the user's phone to access the account, making it harder for hackers to access a user's data. 
\\
\hline
Human
&
EVE Isk Per Hour(Eveiph) is hands down the best tool I've ever used to make isk in New Eden. It is a market helper program that is able to do a great deal of the work that is typically done by a traders spreadsheet. I've used it to go from a 200m/month trading income to 3b/month on my main trading character.

Above you can see the blueprint manufacturing page which is located on the first tab of Eveiph. Here you can see the components required to make an item, the settings for the blueprint, and a brief market analysis of what you can expect to make manufacturing the item and selling it at the market you've selected. You can enter the amount of runs you want to make, the ME and PE of your blueprint and click add to shopping list, and it will be added to a list of items to purchase when you are next at a trade hub.
\\
\hline
Machine
&
So, not only was the speech a thoroughly mediocre diatribe about what he now thinks we should do for the next 45 minutes, but also how much credit we should give to Mumford and Sons for bringing Obama to the campaign trail. Behold:

At the DNC, we drew strength from something even more powerful than the power of words. We drew strength from the power of families in this country. We drew strength from the power of family values. We drew strength from the power of a common purpose--We drew strength from our shared commitment to fighting against everything that undermines our potential in this country and our freedom. It is with that same conviction that we launch this campaign today and we urge every American in America to join us tonight.

To allow the same attempt to succeed in this election.
\\
\hline
Machine
&
The year is twenty-eight, and the boy is Harry, the sixth year at Hogwarts School of Witchcraft and Wizardry. He can't walk without spells covering his feet (or in his case, his feet are so badly burned that he, for practical purposes, can't even walk for that long without them) and he's just starting to feel more secure about things. This is a pretty dull aspect of the book, I'd say. They probably spent way too much time on the fact that he can't use the stick of silver from his wand, despite his friends bewitching all the knives they had.

Harry had been having some difficulty getting to sleep until Hermione pulled him out of his state of near-death-conversation. Thanks to Hermione's meddling, he's gotten some sleep for the past two days. They also learnt a fair amount about getting used to his new surroundings.
\\
\hline
Machine
&
Coincidentally, just a few days after the first tweet came out, a fellow named Kevin McReynolds sent out an interview with GQ to promote their upcoming issue.

McReynolds describes himself as "a conservative Catholic" who "cannot fathom this guy being a real person and should be ashamed that he was able to be elected president."

It's true. If you believe Hillary Clinton gave away 20 percent of the American Uranium to Russia, then you should be ashamed that you voted for Trump. No one should be able to give or receive anything that's not supposed to, so long as they have a warrant. If you've been in a relationship for more than six months with a person who's also convicted of being a felon (or convicted of stealing), that's just stupid, especially as a married man. If you're married to someone convicted of a crime, and they go on their honeymoon with you, that's a felony, not a honeymoon.
\\
\hline
Human
&
CHIP DESIGNER Texas Instruments unveiled a family of system on chip (SoC) processors aimed at automakers today, which are designed for use in self-driving cars.

Named the TDA2x, the SoC family integrates safety features, such as aiding auto designers to create advanced driver assistance systems (ADAS), which in turn help "reduce the number of collisions on the road and enable autonomous driving experiences".

"TDA2x device family combines an optimal mix of high performance, vision analytics, video, graphics and general purpose processing cores in a low power envelope, enabling a broad range of ADAS applications including front camera, surround view and sensor fusion," Texas Instruments said in its release.
\\
\hline
Machine
&
Description

This classic blend of coffee, cream, and sugar is the perfect drink! It is a smooth and creamy coffee with hints of cream and sweet sugar that can be enjoyed even after a full day of work or playing! The sugar provides a wonderful texture to the coffee beans, so that it can be scooped out into a cup.

Available in four flavours: vanilla cream, caramel cream, coffee creme, and chocolate cream.

Note: Coffee can be prepared in less than 120 minutes.
Note: Serves one.
\\
\hline
    \end{tabular}
    \caption{The 10 examples that ``expert" raters were guided through before they were asked to perform the detection task. These are hand-selected to showcase the spectrum of generated text and human-written text.}
    \label{tab:expert_rater_training}
\end{table*}

\end{document}